# A lightweight convolutional neural network for target detection in an edge environment

**Hughes Perreault, PhD[1], Ola Ahmad, PhD[1], Daniel Czuboka[2], Eng.**
[1]Thales Digital Solutions, Montreal Canada
[2]Thales Canada, Defence and Security, Optronics

## ABSTRACT

*This paper proposes a novel, lightweight convolutional neural network (CNN) designed for edge device deployment. The CNN is developed to recognize and classify threats or targets of interest. The model is equipped with a real-time hotspot detection algorithm, which enables it to process information quickly and accurately, resulting in improved inference time of up to 20fps and a detection range of more than 400 meters with typical thermal situational awareness sensors. The proposed CNN has been rigorously tested and validated to demonstrate its high accuracy, sensitivity, and low false positive rate, providing a reliable solution for edge devices to detect potential targets. Additionally, the CNN's lightweight design allows for easy deployment, making it an ideal solution for extending the security bubble for edge devices. The proposed model can be used for a variety of applications, including surveillance, security, and object recognition.*

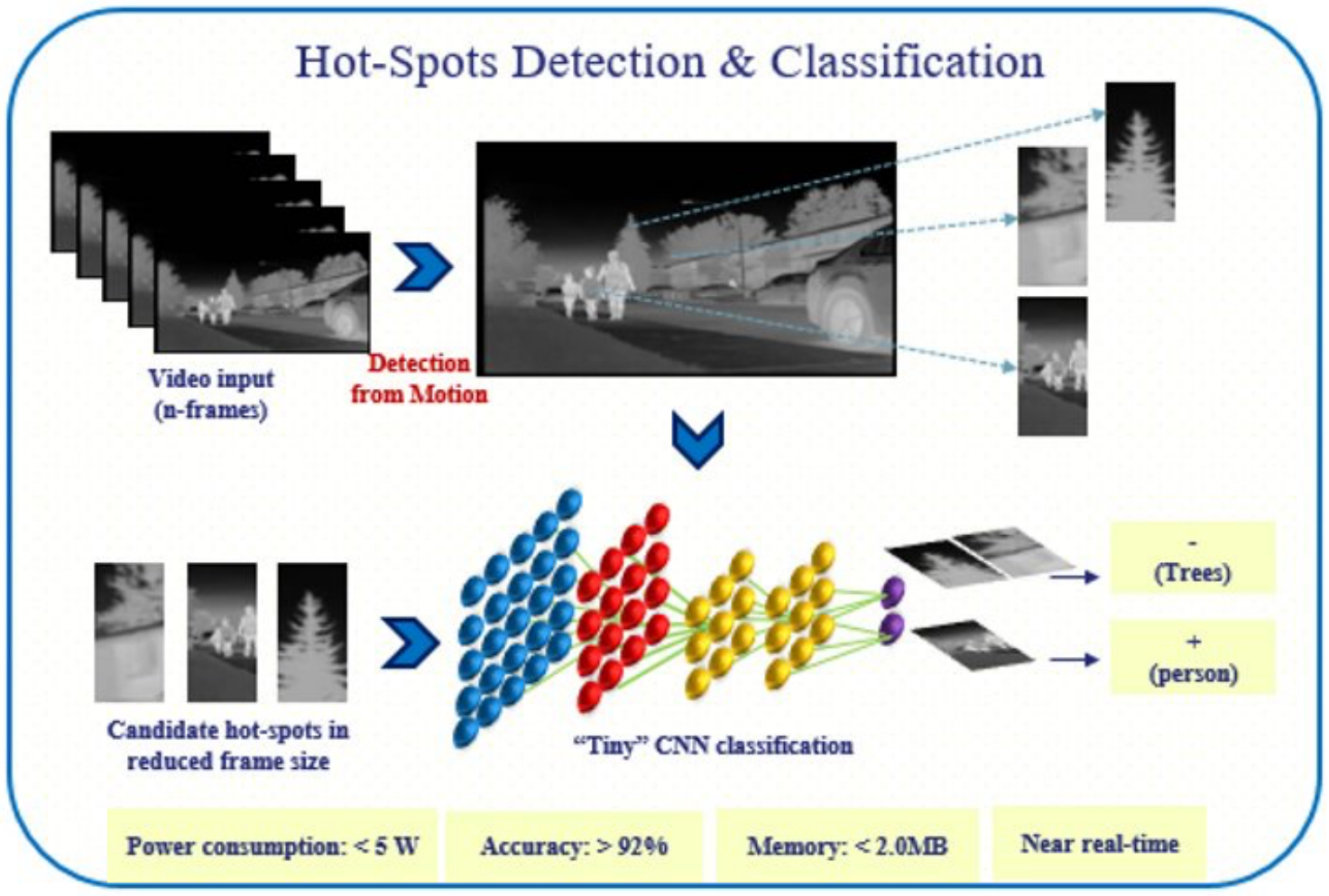

**Figure 1:** An overview of the proposed framework for threats detection and classification.

# 1. INTRODUCTION

Land platform crews find themselves with an ever-increasing workload due to the multitude of sensors and vast quantities of data they generate. Assistive technologies are required to help reduce this cognitive overload particularly in the observation of imagery from both visual and thermal sensors.

Furthermore, the environment around the vehicle can be very busy with people, other vehicles, and potential threats that can come in and out of the scene as the mission proceeds. Human operators and stakeholders need to perceive an event in the present situation, understand its significance and anticipate its change, to make decisions and perform actions. For example, in a combat environment, the crew should be able to detect, recognize and identify persons or objects of interest, comprehend their context, predict their behavior, share the information across a decision network and perform the required actions.

Thermal imaging [1]–[3]is justified here as it is a handy tool for threat and object detection due to its ability to detect objects with elevated temperature, which can indicate the presence of a target of interest. Thermal cameras can pick up on small temperature deltas, which allows for sensing people, vehicles, and animals from a distance day and night. This makes it a great tool for security and military personnel who need to detect threats before they arrive in proximity, maintaining a security bubble around the platform. Contrarily to visible ones, thermal cameras can also sense objects in low light or complete darkness, which further increases their advantage over their more widespread counterpart.

The thermal imagers we are interested in need to be integrated into an edge environment comprising small digital image processing units. This environment allows for collecting the data stream, extracting information, and performing target detection and identification at the edge. Existing target detection and identification algorithms rely on computationally- and memory-intensive models such as deep neural networks. Therefore, it is crucial to leverage edge processing techniques to redesign, compress, deploy, and integrate target detection and classification models on edge.

In this work, we propose a novel framework for target detection in an edge environment (see Fig. 1). The framework comprises two main algorithms: a motion detector and a classifier. The classifier is based on a novel lightweight convolutional neural network (CNN); we call it MiniSQNet. The advantages of the proposed method are threefold. First, it is sensitive to objects smaller than the ones detected by conventional edge CNN models[4], [5]. Second, it has a low latency and can identify objects of interest quickly. The proposed MiniSQNet doesn't need to process the entire (2K or 4K) high-resolution image. It only operates on small regions with high likelihood of a target of interest. Third, it provides accurate localization of the detected targets.

Our method can detect and classify objects or threats with a high degree of accuracy and sensitivity, offering a typical detection range of over 400 meters. The proposed lightweight CNN is a reliable solution for edge devices since it consumes less power (~ 6 Watts) and occupies less than 2 MB memory, which makes it an ideal solution edge environments such as smart sensors.

The rest of the paper is organized as follows: Section 2 provides an overview of related work, Section 3 describes the methodology and the architecture of the proposed model, Section 4 provides experimental results, and Section 5 concludes the paper and provides directions for future work.

## 2. Related Works

In this section, we present an overview of methods related to our work.

**Compact Model design** for light networks refers to the process of creating a deep neural network (DNN) model with a compact architecture that has fewer parameters and requires less computational resources compared to larger models. The goal of small model design is to balance the trade-off between model size and accuracy, such that the smaller model still performs well for the target task.

To achieve this goal, various techniques are employed such as reducing the layers in the model, using lighter operations such as depth-wise separable convolutions, and removing redundant or less impactful features. The specific design choices will depend on the specific task, the available computational resources, and the desired trade-off between model size and accuracy. The most notable works in this field include MobileNet [6]–[8], ShuffleNet [9], [10], SqueezeNet [11] and EfficientNets [12].

Small model design is particularly important for deployment on edge devices, where computational resources are limited, and the models need to be efficient and fast. Additionally, smaller models can be more accessible for training and deployment in low-resource settings.

**Model Compression** is a technique used to reduce the size of deep learning models while maintaining their accuracy. Pruning is the most popular neural network compression technique used in this regard. Pruning is an optimization strategy. It iteratively removes redundant or less critical neurons or weights and retrains the model to compensate for the performance error. It has been applied to a wide range of models, including CNNs, recurrent neural networks, and transformers. Pruning can either be rule-based or learning-based [13], and both approaches are still investigated in the literature. There can be several levels of granularities for pruning [14], fine-grained, vector-level, kernel-level group-level and finally filter-level pruning. Some notable works in this field include [15]–[18].

**Target detection** is a widely researched field in computer vision. Numerous approaches have been proposed for object detection, ranging from traditional methods such as SVM [19] and Haar cascades [20], to more recent deep learning methods such as YOLO [21]–[27] and Faster R-CNN [28]–[30].

In the context of target or object detection, several researchers have applied model pruning to reduce the size of DNN models while maintaining their accuracy. For example, Chen et al. proposed a pruning method for YOLOv3 [31], and Li et al. proposed a pruning method for MobileNet [32]. Our method is different in twofold. First, it performs classification which makes it lighter and faster. Second, it allows detection of distant (or very small) targets of typically more than 400 meters, while baseline methods are limited to 60 meters detection from thermal sensors.

## 3. Methodology

We describe in this section the proposed framework shown in Fig. 1.

### 3.1. Motion detector

We first use a motion detector to extract candidate targets from each frame of data stream. In this work, we assume that targets of interest are moving objects. For instance, persons performing suspicious activities or moving in specific areas, vehicles moving in the scene with respect to the camera, etc. In this work we focus on three types of targets, moving persons, vehicles, and drones. The motion detector is a classical algorithm that detects local changes in each frame and indicates the pixels containing the maximum likelihood ratio of change. This classical algorithm uses standard background subtraction techniques to isolate the hot-spot location. If the position of the hot spot with respect to the scene changes slightly for three successive frames, then we determine that it is in fact a hot moving object (potential threat). We use these pixels to generate patch images including ROI (Regions-Of-Interest) for the classifier.

### 3.2. Classifier (MiniSQNet)

We propose a novel edge algorithm, a lightweight CNN called MiniSQNet, to predict if the ROI image is an actual target of interest. Our method relies on DNN compression techniques. More specifically, we first adopt a compact neural network architecture called SqueezeNet (SQNet [11]). Although this model is tailored to tackle hardware resource constraints, it is still considered as relatively complex for embedded applications with limited power and memory budgets. We propose to modify the design of the original SQNet making it at least 2x smaller while providing competitive classification accuracy.

To do so, we combine two compression techniques: pruning and knowledge distillation. For pruning, two common approaches exist greedy pruning, and one-shot pruning. In greedy pruning, the pruning is done in multiple phases where each layer is processed one at a time followed by a retraining. This technique is very time consuming and requires costly training resources. One-shot-pruning on the other hand, removes all the filters of the DNN retrain once, which is much more efficient. However, it requires to choose beforehand the pruning threshold. In our approach, we made the pruning threshold as a hyperparameter that can be tuned to balance compression and performance tradeoffs.

**Table 1.** Validation of our compression strategy.

| Model | Top-1 acc. | Top-5 acc. | #Params | Memory size (MB) |
|---|---|---|---|---|
| SQNet (baseline) | 0.72 | 0.96 | 740,554 | 4.2 |
| Pruned (70%) | 0.52 | 0.88 | 221.774 | 0.993 |
| +distilled | 0.68 | 0.95 | 221774 | 0.993 |
| +quantized (8bits) | 0.67 | 0.94 | 223014 | 0.942 |

Despite that pruning enables light and compressed models, it makes models overfit fast and affect its generalization. To mitigate this issue, we combine pruning with knowledge distillation technique [33]. Its goal is to transfer knowledge from a teacher model, often a well-trained deeper one to a student model, the pruned one in our case. In our experiments, we used the original SQNet as a teacher model.

# 4. Implementation details

## 4.1. Datasets

Annotated thermal datasets are hard to come by. To train our model, we used both open-source and proprietary (in-house) datasets. To train our drone model, we used the Svanstrom [34] dataset for object detection and generate the ROIs using the drone bounding boxes. For pedestrian data, we used CVC-14 [35], FLIR [36] and LSFIR [37]. Examples of these datasets can be viewed in Fig. 2. The vehicles training data are collected using our thermal cameras. Each dataset has been split into a training, validation, and testing sets, and combined to create the final sets.

## 4.2. Training and inference details

**Training details.** We used the training set created from open-source infrared data described in Section 4.1 to train the lightweight CNN. Before the compression phase, we transferred the SQNet (trained on visible images) to our dataset. We keep a copy of this model as we will use it further to guide the training of the smaller model. The training procedure is composed of two phases.

In the first training phase, we implement one-shot pruning and retrain the pruned model (MiniSQNet) for 25 epochs. We chose the pedestrian dataset during this phase since it is the largest one, leading to better and unbiased optimization of the architecture. At the end of these epochs, we distill the knowledge in SQNet into MiniSQNet by adding to the classification loss a distillation loss as described in as described in [33], and retrain for another 10 epochs. At the end, we obtain a light well-optimized model ready to be transferred and scaled to another classes and datasets.

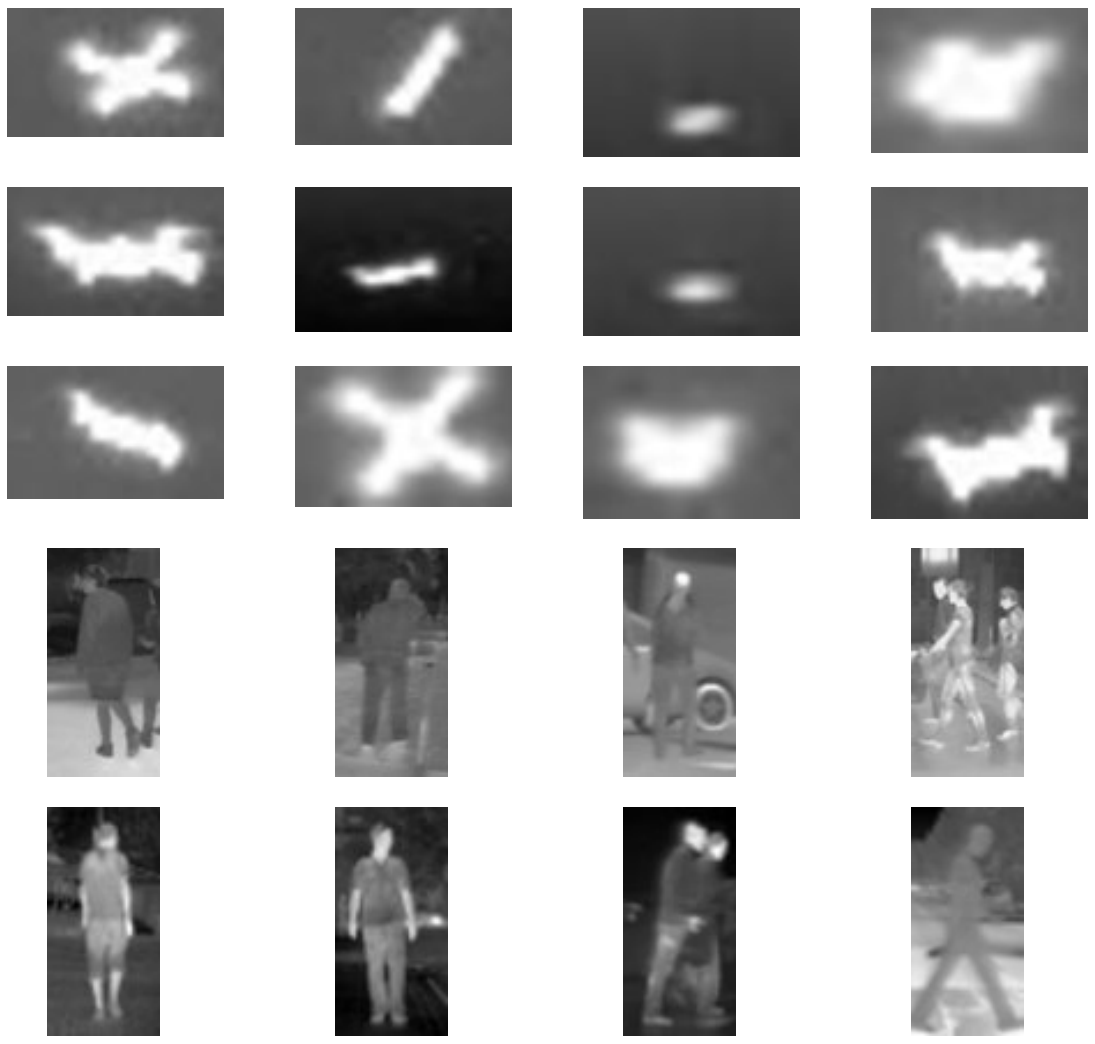

**Figure 2:** Samples from our open-source datasets

The second training phase is model fine-tuning, while we added multi-heads to the classifier to account for the other classes, vehicles, and drones. The finetuning phase uses only the cross-entropy loss as no further compression is implemented at this stage.

**Inference details**. We first apply a fast motion detection algorithm to implement the target detection pipeline, as mentioned in Section 3.1. The algorithm is composed of background subtraction, followed by a temporal Hessian detector [39]. This method is susceptible to local intensity changes at each pixel and effectively captures moving targets emitting radiation.

Once the candidate pixels are detected, patches centered at these pixels are cropped and extracted to be analyzed. The patches are then resized and fed into the trained lightweight CNN to classify each ROI as a target or not a target of interest.

# 5. Experiments

## 5.1. Evaluating the proposed compression strategy

We tested different pruning thresholds (0.6, 0.7, and 0.8). We found that aggressive pruning at 0.8 destroyed the topology in specific layers and made the model unable to train. On the other hand, setting the threshold to lower values doesn't provide an advantage in terms of compression. Accordingly, we fixed the pruning threshold to 0.7 in all our experiments. That means we kept only 30% of the filters, where the score of each filter in each layer is related to the activation strength.

In Table 1. we report the evaluation results on the CIFAR10 [8] dataset and compare them to the baseline (SQNet). As can be seen, pruning the model leads to a slight performance drop even after retraining, which is an indicator of overfitting on the test set. Using Knowledge distillation compensates for the performance drop and increases the accuracy to a competitive value compared to the baseline. We also applied a post-training quantization to analyze its effect on performance. As the results show, quantizing the weights from 32bits (floating-point) to 8bits after training did not affect the performance and allowed to remove 50 bytes of model size further.

**Table 2**. Results of the proposed target detection model on test data.

| Edge CNN Models | Threat | Distance range (m) | F1-Score | FPR (%) | Image res. (pixels) | Memory size (MB) | Power consumptions |
|---|---|---|---|---|---|---|---|
| **One-head** | person | [25 – 300] | 0.96 | 2 | < 4000 | 0.993 | < 6 watts |
| | drone | NA | 0.99 | 0.35 | | | |
| **Multi-heads** | All | [25 – 450+] | 0.79 | 2.6 | | 1.017 | NA |
| **Baseline (YOLOV3)** | person only | [25-60] | 0.85 | NA | > 200K (Full) | > 20 | NA |

## *5.2. Evaluating the lightweight CNN performance on target detection*

Our experiments consisted of fine-tuning the classification network for each use case, using a validation set to determine to perform early stopping based on the validation loss value. The classification accuracy was then computed on a separate testing set for the image patches. In a later stage, the entire pipeline (including the motion detector) is visually validated on testing videos. To demonstrate the versatility of our model, three experiments were performed. One for persons, one for drones, and a final one for differentiating between vehicles, drones and vehicles threats simultaneously. We report the results of our experiments in table 2, and compare it to a YOLOv3 baseline.

Our tiny model shows great performance, being approximately 1 MB in size, and the entire pipeline consuming less than 6 watts of power. The YOLOv3 baseline is more than 20 times that size in comparison, and does not perform as well, especially for small or distant threats. This excellent efficiency comes from the fact that due to our candidate selection with the hotspot algorithm, the whole image never needs to get processed, only the relevant parts.

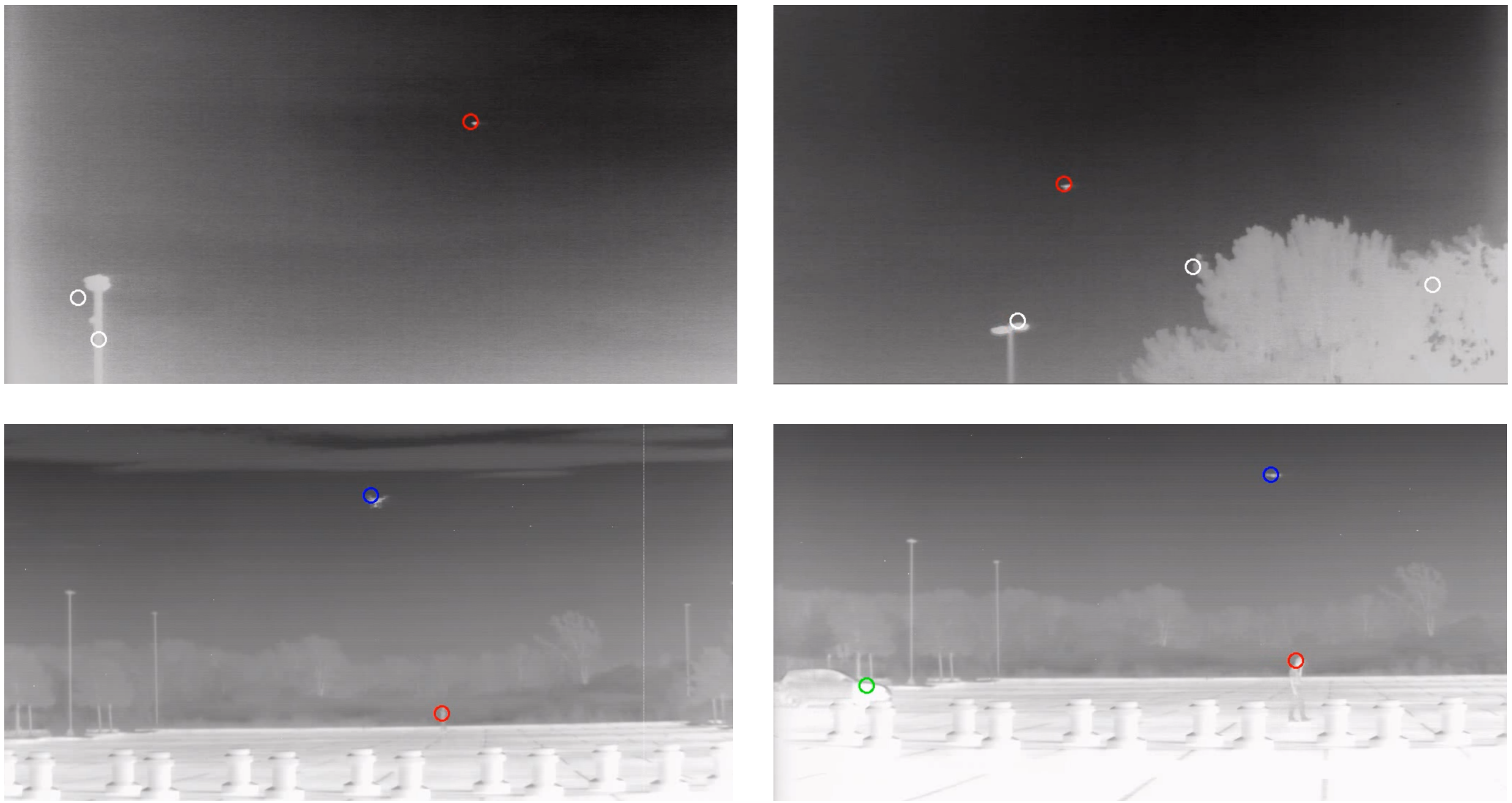

**Figure 3:** Qualitative results of our method. Top left and top right: a drone is detected (red) while hotspots (white) are classified as non-threats. Bottom left: a drone (blue) and a person (red) are detected. Bottom right: a drone (blue), a person (red) and a vehicle (green) are detected. Better seen in color.

In terms of accuracy, we obtain excellent F1-Score and false positive rate (FPR) for one class threat detection, and slightly lower F1-score and FPR for three class detection, which is understandable due the occasional misclassification of a threat for another type of threat. Some qualitative results are shown in figure 3, where we can see different types of threats being identified, and hotspots being identified as non-threats.

## 6. Conclusion

The lightweight CNN proposed in this paper is a promising approach to effectively operate on constrained SWaP (Size-Weight-and-Power consumption) hardware and edge environments. The proposed model achieved impressive detection results at a low computational cost, making it a viable solution for real-world applications. It can be addition, the proposed model can be easily adapted to different environments and datasets, thus providing a generalizable model for the target (and threat) detection. We emphasize that the goal of the lightweight CNN proposed in this work is to spot potential threats and sense objects of interest regardless of the lightning conditions as soon as possible (in real-time) and at the farthest distances by relying on sensing distant moving hot spots in the scene. (i.e., using infrared sensors). Once the target is detected and localized, a higher resolution visible (RGB) or SWIR camera can be used to point at and zoom in on the target, capturing additional details that aid in threat analysis. Our experiments highlight the potential of the proposed thermal target detection framework in increasing the security bubble around edge devices and their carriers. Future work might include better using the temporal aspect (i.e., using video information instead of static images) to track targets and making the model faster, lighter, and more robust.